\documentclass[10pt,twocolumn,letterpaper]{article}

\usepackage{cvpr}              

\usepackage{graphicx}
\usepackage{amsmath}
\usepackage{amssymb}
\usepackage{booktabs}
\usepackage{tabularx}
\usepackage{boldline}

\usepackage[pagebackref,breaklinks,colorlinks]{hyperref}

\usepackage[capitalize]{cleveref}
\crefname{section}{Sec.}{Secs.}
\Crefname{section}{Section}{Sections}
\Crefname{table}{Table}{Tables}
\crefname{table}{Tab.}{Tabs.}

\def\cvprPaperID{2369} 
\def\confName{CVPR}
\def\confYear{2023}

\begin{document}


\title{Scene-aware Egocentric 3D Human Pose Estimation}

\author{Jian Wang\textsuperscript{1,2}~~~~~~Diogo Luvizon\textsuperscript{1,2}~~~~~~Weipeng Xu\textsuperscript{3}~~~~~~Lingjie Liu\textsuperscript{1,2}\\
Kripasindhu Sarkar\textsuperscript{4}~~~~~~Christian Theobalt\textsuperscript{1,2}\\
	\textsuperscript{1}MPI Informatics~~~~~\textsuperscript{2}Saarland Informatics Campus~~~~~\textsuperscript{3}Meta Reality Labs~~~~~\textsuperscript{4}Google\\
	{\tt\small \{jianwang,dluvizon,lliu,theobalt\}@mpi-inf.mpg.de~~xuweipeng@fb.com~~krsarkar@google.com}
}
\maketitle

\begin{abstract}
Egocentric 3D human pose estimation with a single head-mounted fisheye camera has recently attracted attention due to its numerous applications in virtual and augmented reality. Existing methods still struggle in challenging poses where the human body is highly occluded or is closely interacting with the scene. To address this issue, we propose a scene-aware egocentric pose estimation method that guides the prediction of the egocentric pose with scene constraints. To this end, we propose an egocentric depth estimation network to predict the scene depth map from a wide-view egocentric fisheye camera while mitigating the occlusion of the human body with a depth-inpainting network. Next, we propose a scene-aware pose estimation network that projects the 2D image features and estimated depth map of the scene into a voxel space and regresses the 3D pose with a V2V network. The voxel-based feature representation provides the direct geometric connection between 2D image features and scene geometry, and further facilitates the V2V network to constrain the predicted pose based on the estimated scene geometry. To enable the training of the aforementioned networks, we also generated a synthetic dataset, called EgoGTA, and an in-the-wild dataset based on EgoPW, called EgoPW-Scene. The experimental results of our new evaluation sequences show that the predicted 3D egocentric poses are accurate and physically plausible in terms of human-scene interaction, demonstrating that our method outperforms the state-of-the-art methods both quantitatively and qualitatively.
\end{abstract}

\section{Introduction} \label{sec:intro}

Egocentric 3D human pose estimation with head- or body-mounted cameras is extensively researched recently because it allows capturing the person moving around in a large space, while the traditional pose estimation methods can only record in a fixed volume. With this advantage, the egocentric pose estimation methods show great potential in various applications, including the xR technologies and mobile interaction applications.

\begin{figure}
\centering
\includegraphics[width=0.96\linewidth]{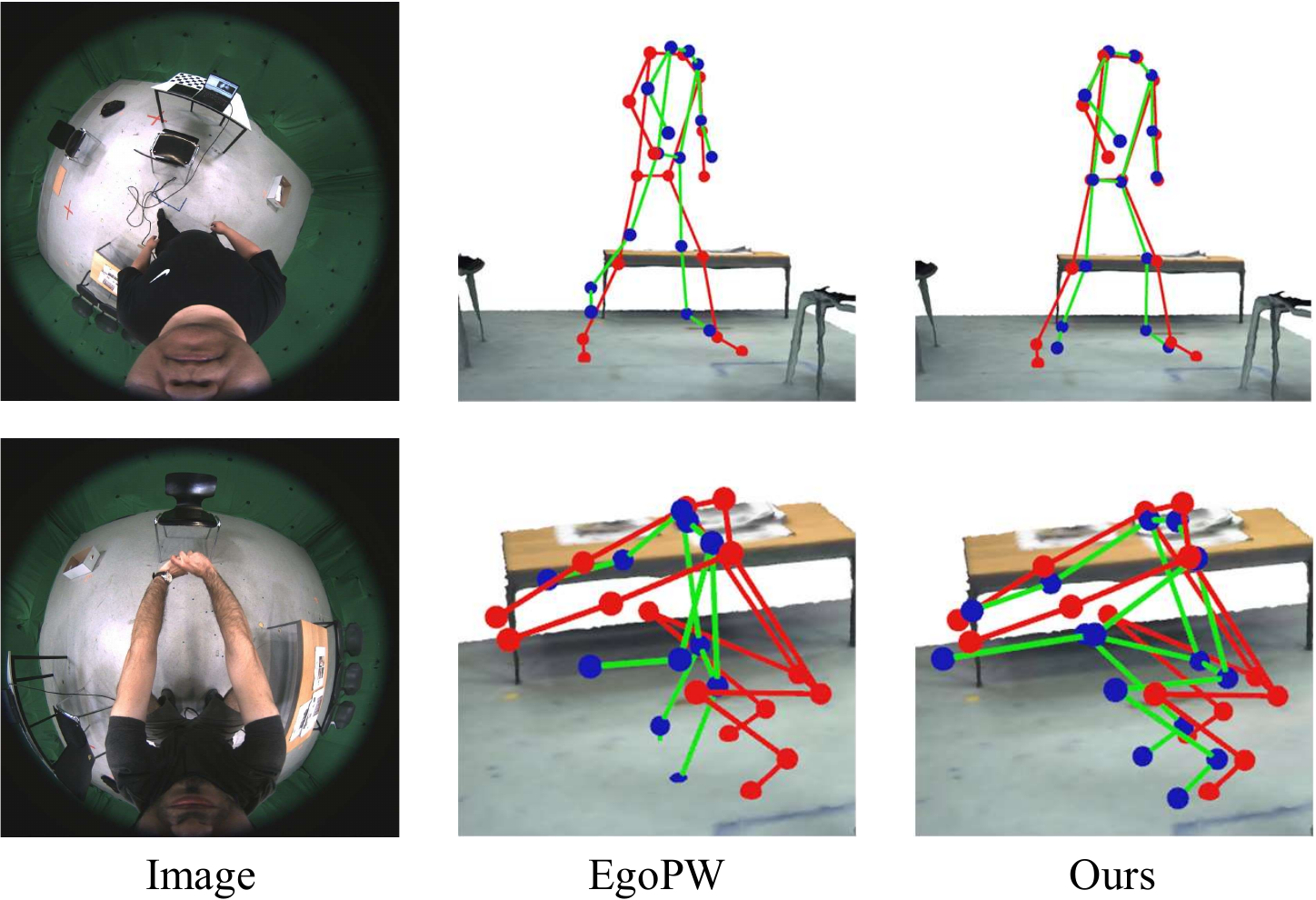}
\caption{Previous egocentric pose estimation methods like EgoPW predict body poses that may suffer from body floating issue (the first row) or body-environment penetration issue (the second row). Our method predicts accurate and plausible poses complying with the scene constraints. The red skeletons are the ground truth poses and the green skeletons are the predicted poses. 
}
\label{fig:teaser}
\end{figure}

In this work, we estimated the full 3D body pose from a single head-mounted fisheye camera. A number of works have been proposed, including Mo$^2$Cap$^2$~\cite{DBLP:journals/tvcg/XuCZRFST19}, $x$R-egopose~\cite{DBLP:conf/iccv/TomePAB19}, Global-EgoMocap~\cite{wang2021estimating}, and EgoPW~\cite{Wang_2022_CVPR}. These methods have made significant progress in estimating egocentric poses. However, when taking account of the interaction between the human body and the surrounding environment, they still suffer from artifacts that contrast the physics plausibility, including body-environment penetrations or body floating (see the EgoPW results in Fig.~\ref{fig:teaser}), which is 
mostly ascribed to the ambiguity caused by the self-occluded and highly distorted human body in the egocentric view. This problem will render restrictions on subsequent applications including action recognition, human-object interaction recognition, and motion forecasting.

To address this issue, we propose a scene-aware pose estimation framework that leverages the scene context to constrain the prediction of an egocentric pose. This framework produces accurate and physically plausible 3D human body poses from a single egocentric image, as illustrated in Fig.~\ref{fig:teaser}. 
Thanks to the wide-view fisheye camera mounted on the head, the scene context can be easily obtained even with only one egocentric image. To this end, we train an egocentric depth estimator to predict the depth map of the surrounding scene. In order to mitigate the occlusion caused by the human body, we predict the depth map including the visible human and leverage a depth-inpainting network to recover the depth behind the human body.

Next, we combine the projected 2D pose features and scene depth in a common voxel space and regress the 3D body pose heatmaps with a V2V network~\cite{moon2018v2v}. The 3D voxel representation projects the 2D poses and depth information from the distorted fisheye camera space to the canonical space, and further provides direct geometric connection between 2D image features and 3D scene geometry. 
This aggregation of 2D image features and 3D scene geometry facilitates the V2V network to learn the relative position and potential interactions between the human body joints and the surrounding environment and further enables the prediction of plausible poses under the scene constraints.
%


Since no available dataset can be used for train these networks, we proposed EgoGTA, a synthetic dataset based on the motion sequences of GTA-IM~\cite{caoHMP2020}, and EgoPW-Scene, an in-the-wild dataset based on EgoPW~\cite{Wang_2022_CVPR}. Both of the datasets contain body pose labels and scene depth map labels for each egocentric frame.

To better evaluate the relationship between estimated egocentric pose and scene geometry, we collected a new test dataset containing ground truth joint positions in the egocentric view.
The evaluation results on the new dataset, along with results on datasets in Wang~\etal~\cite{wang2021estimating} and Mo$^2$Cap$^2$~\cite{DBLP:journals/tvcg/XuCZRFST19} demonstrate that our method significantly outperforms existing methods both quantitatively and qualitatively.
We also qualitatively evaluate our method on in-the-wild images. 
The predicted 3D poses are accurate and plausible even in challenging real-world scenes. 
To summarize, our contributions are listed as follows:
\begin{itemize}
    \item The first scene-aware egocentric human pose estimation framework that predicts accurate and plausible egocentric pose with the awareness of scene context;
    \item Synthetic and in-the-wild egocentric datasets containing egocentric pose labels and scene geometry labels;\footnote{Datasets are released in our \href{https://people.mpi-inf.mpg.de/~jianwang/projects/sceneego/}{project page}. Meta did not access or process the data and is not involved in the dataset release.}
    \item A new depth estimation and inpainting networks to predict the scene depth map behind the human body;
    \item By leveraging a voxel-based representation of body pose features and scene geometry jointly, our method outperforms the previous approaches and generates plausible poses considering the scene context.
\end{itemize}

%
\section{Related Work} \label{sec:relatedworks}
\begin{figure*}[h]
\centering
\includegraphics[width=0.96\linewidth]{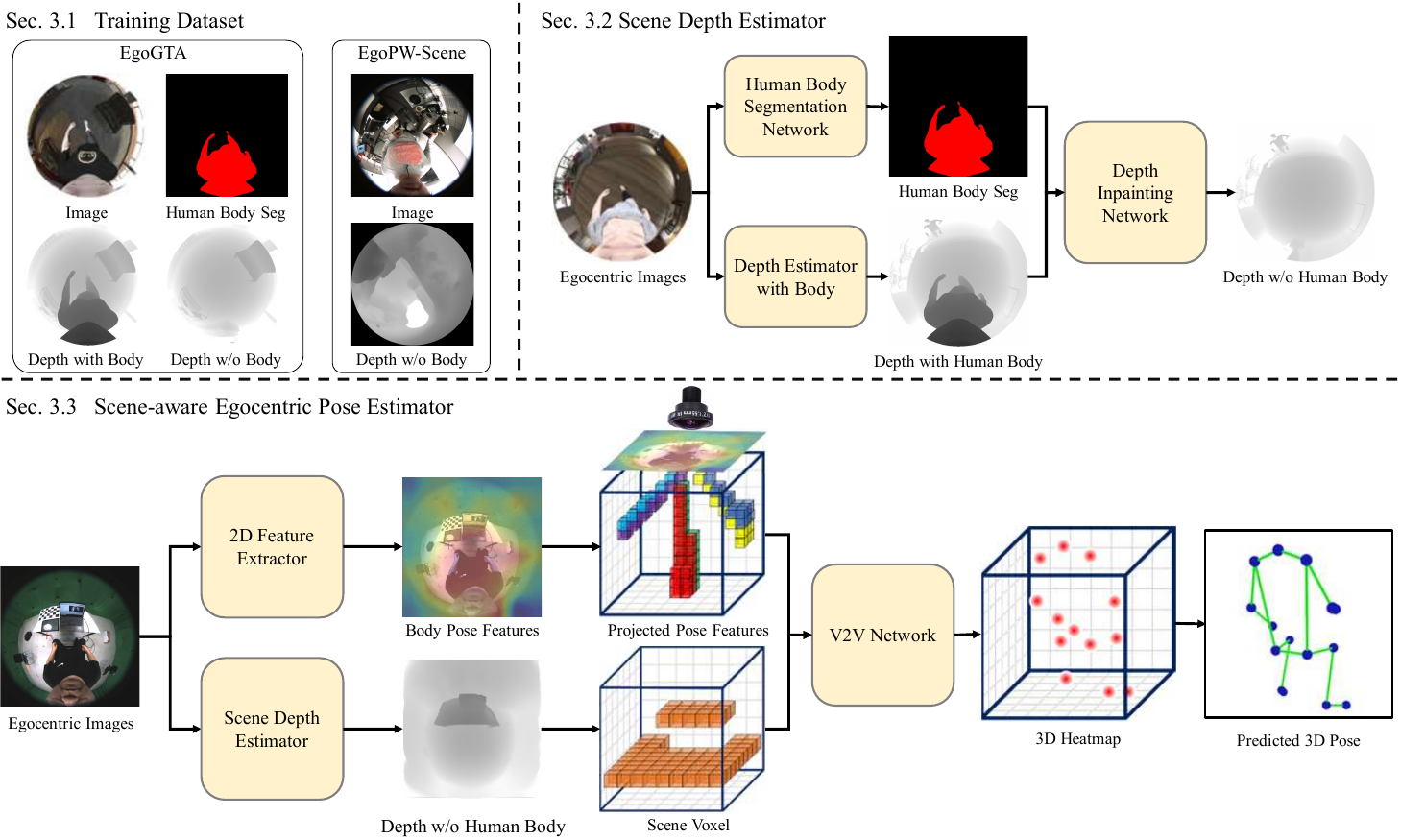}
\caption{Overview of our method.
We first render synthetic training dataset EgoGTA and in-the-wild training dataset EgoPW-Scene. Both datasets contain egocentric depth maps for subsequent training process (Sec.~\ref{method:dataset}). Next, we train an egocentric scene depth estimator that predicts a depth map without the human body and a depth inpainting network (Sec.~\ref{method:depth}). Finally, we combine the 2D body pose features and scene depth map into a common voxel space. The 3D body pose heatmaps are regressed from the voxel space with a V2V network and the final pose prediction is obtained with soft-argmax (Sec.~\ref{method:pose}).
}
\label{fig:framework}
\end{figure*}

\subsection{Egocentric 3D Full Body Pose Estimation}
%
%
Inspired by the new applications in augmented reality and by the limitations of traditional motion capture systems, Rhodin \etal~\cite{DBLP:journals/tog/RhodinRCISSST16} proposed the first egocentric motion capture system based on a pair of fisheye cameras.
The following methods proposed new architectures~\cite{cha2018fullycapture, EgoGlass_3DV_2021} and new datasets~\cite{EgoGlass_3DV_2021, hakada2022unrealego} for stereo egocentric pose estimation.
However, a stereo camera setup implies additional computation complexity and extra energy consumption, which is critical for low-power head-mounted devices, which are the main target applications.
\par
The single head-mounted fisheye camera setup was first proposed by Xu~\etal~\cite{DBLP:journals/tvcg/XuCZRFST19}, who also introduced a two-stream CNN to cope with the low resolution of regions far from the fisheye camera, \ie, one branch for the full body and one branch for predicting the lower body joints from a zoom-in image.
Tome \etal~\cite{DBLP:conf/iccv/TomePAB19} proposed an encoder-decoder architecture to model the high uncertainty caused by severe occlusions present in this setup and
Wang~\etal~\cite{wang2021estimating} leverages motion capture data to learn a human motion prior, which is applied in an optimization method to obtain temporally stable poses for training and egocentric pose predictor.
Another challenge is the strong image distortion caused by the fisheye lens, which can be mitigated with automatic camera calibration~\cite{zhang2021automatic}.
Other setups consider the camera facing forward and try to synthesize plausible human motion given only scene image evidences~\cite{luo2021dynamics} or partially visible body parts~\cite{jiang2021egocentric, Hwang2020}.
%
%
In our work, we do not consider this setup since in many poses only a few body extremities are visible in the image.
\par
Obtaining real data in the egocentric setup is a time consuming process, therefore, many approaches rely on synthetic data for training~\cite{DBLP:conf/iccv/TomePAB19, jiang2021egocentric}. A recent method has shown that the existing gap between synthetic and real data can be mitigated by domain adaptation techniques~\cite{Wang_2022_CVPR}, but this approach still requires real data for the weak supervision part.
Differently, our method has a simpler training strategy, and we reduce the gap between synthetic and real data by providing additional pseudo ground-truth scene labels for real sequences, including indoor and outdoor scenes.

\subsection{Voxel Representation for Body Pose Estimation}
Volumetric voxel representations have been extensively used with multiple view setups for the estimation of single~\cite{Shimosaka_ICRA_2009, trumble_eccv_2018} and multiple human poses~\cite{voxelpose_ECCV_2020, Zhang_TPAMI_2022, fastervoxelpose, wang2021mvp}, and for hand pose estimation~\cite{Huang2018StructureAware3H, Moon_2018_CVPR_V2V, HandVoxNet2020, HandVoxNet++2021} from depth maps input. 
Considering a single image as input, Pavlakos \etal~\cite{pavlakos17volumetric} proposed a coarse-to-fine approach to lift from 2D heatmaps to a voxel representation for 3D human pose estimation. Iskakov~\cite{iskakov2019learnable} later proposed a learnable triangulation method in voxel space that can generalize to single or multiple views, achieving less than $2$~cm error in a controlled multi-view data.
Despite the success of voxel representation for human and hand pose estimation from an external camera, this technique has not yet been explored for the interaction between human body and scene. In our work, we show the advantages of this representation for egocentric human pose estimation, especially when considering human-scene interaction.

\subsection{Scene-aware Human Pose Estimation}
In recent years, several approaches have been proposed to predict the pose of humans considering environmental and physical constraints from RGB~\cite{PhysCapTOG2020, yi2022mover, pavlakos2022sitcoms3D} and inertial measurement units (IMU)~\cite{HPS, yi2022physical}.
Some methods assume a simplified environment, such as a planar ground floor, to enforce a temporal sequence that is physically consistent with the universal law of gravity by assuming known camera poses~\cite{PhysCapTOG2020, PhysAwareTOG2021} or by tracking an object in the scene following a free flight trajectory~\cite{Dabral_2021_ICCV}.
Other approaches assume that the scene is given as input, either as a 3D reconstruction~\cite{HPS, shimada2022hulc} or as geometric primitives~\cite{Yu:2021:MovingCam}, whose positions can be refined in the optimization process.
Bhatnagar \etal~\cite{bhatnagar2022behave} proposed a method and dataset for human-object interactions. Taking into account the interaction between humans and furniture, holistic methods are able to estimate the position of humans and specific objects in the scene under the assumption of a planar floor~\cite{weng2021holistic, chen2019holistic++, yi2022mover}, or even to estimate deformations in known objects based on human poses~\cite{Li_3DV2022}.
Contrary to the previous work, we make no strong assumptions about the objects and ground floor in the scene, but instead propose a method that learns to estimate the background scene geometry from a fisheye camera and explores the correlation between the human body and scene directly from egocentric data.

\section{Method} \label{sec:method}
We propose a new method for predicting accurate egocentric body pose by leveraging the estimated scene geometry. An overview of our method is shown in Fig.~\ref{fig:framework}. In order to train the scene-aware network, we first generate a synthetic dataset based on the GTA-IM dataset~\cite{caoHMP2020}, called EgoGTA, and an in-the-wild dataset based on the EgoPW dataset~\cite{Wang_2022_CVPR}, called EgoPW-Scene (Sec.~\ref{method:dataset}).
Next, we train a depth estimator to estimate the geometry of the surrounding scene and introduce the depth-inpainting network that estimates the depth behind the human body (Sec.~\ref{method:depth}).
Finally, we combine 2D features and scene geometry in a common voxel space and predict the egocentric pose with a V2V network~\cite{moon2018v2v} (Sec.~\ref{method:pose}). 
%

\subsection{Training Dataset}\label{method:dataset}
%
Although many training datasets for egocentric pose estimation~\cite{DBLP:journals/tvcg/XuCZRFST19, DBLP:conf/iccv/TomePAB19, Wang_2022_CVPR} have been proposed, they cannot yet train the scene-aware egocentric pose estimation network due to the lack of scene geometry information. To solve this, we introduce the EgoGTA dataset and EgoPW-Scene dataset (both will be made publicly available). Both datasets contain pose labels and depth maps of the scene for each egocentric frame, facilitating our training process. We show examples from both datasets, as illustrated in Fig.~\ref{fig:framework}. 


\subsubsection{EgoGTA Dataset} 
In order to obtain precise ground truth human pose and scene geometry for training, we devise a new synthetic egocentric dataset based on GTA-IM~\cite{caoHMP2020}, which contains various daily motions and ground truth scene geometry. For this, we first fit the SMPL-X model on the 3D joint trajectories from GTA-IM. Next, we attach a virtual fisheye camera to the forehead of the SMPL-X model and render the images, semantic labels, and depth map of the scene with and without the human body.
In total, we obtained $320$ K frames in  $101$ different sequences, each with a different human body texture. Here, we denote the EgoGTA dataset $\mathbb{S}_G = \{I_G, S_G, D^B_G, D^S_G, P_G\}$, including synthetic images $I_G$ and their corresponding human body segmentation maps $S_G$, depth map with human body $D^B_G$, depth map of the scene without human body $D^S_G$, and egocentric pose labels $P_G$.

\subsubsection{EgoPW-Scene Dataset}
Since we want to generalize to data captured with a real head-mounted camera, we also extended the EgoPW~\cite{Wang_2022_CVPR} training dataset.
For this, we first reconstruct the scene geometry from the egocentric image sequences of the EgoPW training dataset with a Structure-from-Motion (SfM) algorithm~\cite{hartley2003multiple}. This step provides a dense reconstruction of the background scene. The global scale of the reconstruction is recovered from known objects present in the sequences, such as laptops and chairs. We further render the depth maps of the scene in the egocentric perspective based on the reconstructed geometry. Our EgoPW-Scene dataset contains $92$ K frames in total, which are distributed in $30$ sequences performed by $5$ actors. The number of frames in the EgoPW-Scene dataset is less than EgoPW dataset since SfM fails on some sequences. Here, we denote the EgoPW-Scene dataset $\mathbb{S}_E = \{I_E, D^S_E, P_E\}$, including in-the-wild images $I_E$ and their corresponding depth map of the scene without human body $D^S_E$, and egocentric pose labels $P_E$.

\subsection{Scene Depth Estimator}\label{method:depth}

In this section, we propose a depth estimation method to capture the scene geometry information in the egocentric perspective. 
Available depth estimation methods~\cite{fu2018deep,lee2019big,hu2019revisiting} can only generate depth maps with the human body, but are not able to infer the depth information behind the human, \ie, the background scene depth.
However, the area occluded by the human body, \eg the areas of foot contact, are crucial for generating plausible poses, as demonstrated in Sec.~\ref{experiments:ablation}.
To predict the depth map of the scene behind the human body, we adopt a two-step approach. More specifically, we first estimate the depth map including the human body and the semantic segmentation of the human with two separated models. Then, we use a depth inpainting network to recover the depth behind the human body. This two-step strategy is necessary because the human visual evidences in the RGB images are too strong to be ignored by the depth estimator, therefore, it is easier to train the scene depth estimation as separated tasks.

We first train the depth estimator network $\mathcal{D}$, which takes as input a single egocentric image $I$ and predicts the depth map with human body $\hat{D}^B$.
The network architecture of $\mathcal{D}$ is the same as Hu~\etal~\cite{hu2019revisiting}'s work. To minimize the influence of the domain gap between synthetic and real data, the network is initially trained on the NYU-Depth V2 dataset~\cite{Silberman:ECCV12} following \cite{hu2019revisiting}, and further fine-tuned on the EgoGTA dataset.

Next, we train the segmentation network $\mathcal{S}$ for segmenting the human body. The network takes the egocentric image $I$ as input and predicts the segmentation mask for the human body $\hat{S}$ as output. Following Yuan~\etal~\cite{YuanCW19}, we use HRNet as our segmentation network. Similarly, to reduce the domain gap, we pretrain the network on the LIP dataset~\cite{gong2017look} and finetune the model on the EgoGTA dataset. We do not train network $\mathcal{D}$ and $\mathcal{S}$ on the EgoPW-Scene dataset since it lacks the ground truth segmentation maps and depth maps with the human body.

Finally, we propose a depth inpainting network $\mathcal{G}$ for generating the final depth map of the scene without human body. We first generate the masked depth map $~{\hat{D}^M=(1-\hat{S})\odot\hat{D}^B}$, which is a Hadamard product between the background segmentation and the depth map with human body. Then, the masked depth map $\hat{D}^M$ and the segmentation mask $\hat{S}$ are fed into the inpainting network $\mathcal{G}$, which predicts the final depth map $\hat{D}^S$. We train the inpainting network $\mathcal{G}$ and finetune the depth estimation network $\mathcal{D}$ on both the EgoGTA and EgoPW-Scene datasets.
During training, we penalize the differences between the predicted depth maps and the ground truth depth of the background scene with $L^S$ and also keep the depth map consistent in the non-human body regions with $L^C$. 
%
Specifically, the loss function is defined as follows: 
\begin{equation}
\begin{split}
    L &= \lambda^SL^S + \lambda^CL^C, \quad \text{with} \\
    L^S &= \left\Vert \hat{D}^S_G - D^S_G \right\Vert_2^2 + \left\Vert \hat{D}^S_E - D^S_E \right\Vert_2^2, \quad \text{and} \\
    L^C &= \left\Vert (\hat{D}^S_G - \hat{D}^B_G)(1 - \hat{S}_G) \right\Vert_2^2 \\
    &+ \left\Vert (\hat{D}^S_E - \hat{D}^B_E)(1 - \hat{S}_E) \right\Vert_2^2,
\end{split}
\end{equation}
where
\begin{equation}
\begin{split}
    \hat{D}^S_G &= \mathcal{G}(\hat{D}^M_G, \hat{S}_G);\quad \hat{D}^S_E = \mathcal{G}(\hat{D}^M_E, \hat{S}_E);\\
    \hat{D}^B_G &= \mathcal{D}(I_G);\quad \quad \quad \, \hat{D}^B_E = \mathcal{D}(I_E); \\
    \hat{S}_G &= \mathcal{S}(I_G);\quad \quad \quad \; \; \hat{S}_E = \mathcal{S}(I_E),
\end{split}
\end{equation}
and $\lambda^S$ and $\lambda^C$ are the weights of the loss terms. 


\subsection{Scene-aware Egocentric Pose Estimator}\label{method:pose}

In this section, we introduce our scene-aware egocentric pose estimator.
We rely on the prior that human bodies are mostly in contact with the scene. However, estimating the contact explicitly from a single egocentric image is very challenging. Therefore, we rely on a data-driven approach by learning a model that predicts a plausible 3D pose given the estimated scene geometry and features extracted from the input image.
To achieve this goal, we first leverage the EgoPW~\cite{Wang_2022_CVPR} body joints heatmap estimator to extract 2D body pose features $F$ and use the scene depth estimator from Sec.~\ref{method:depth} to estimate the depth map of the scene without human body $\hat{D}^S$. Afterwards, we project the body pose features and depth map into a 3D volumetric space considering the fisheye camera projection model. After obtaining the volumetric representation of human body features $V_{\text{body}}$ and scene depth $V_{\text{scene}}$, the 3D body pose $\hat{P}$ is predicted from the volumetric representation with a V2V network~\cite{moon2018v2v}. 

Lifting the image features and depth maps to a 3D representation allows getting more plausible results, as inconsistent  joint predictions can be behind the volumetric scene $V_{\text{scene}}$ (pose-scene penetration) or spatially isolated from the voxelized scene geometry (pose floating), so they can be easily identified and adjusted by the volumetric convolutional network. 

\subsubsection{Scene and Body Encoding as a 3D Volume}

In order to create the volumetric space, we first create a 3D bounding box around the person in the egocentric camera coordinate system of size $L \times L \times L$, where $L$ denotes the length of the side of the bounding box in meters. The egocentric camera is placed at the center-top of the 3D bounding box so that the vertices of the bounding boxes are: $(\pm L/2, \pm L/2, 0)$ and $(\pm L/2, \pm L/2, L)$ under the egocentric camera coordinate system. Next, we discretize the bounding box by a volumetric cube $V \in R^{N,N,N,3}$. Each voxel $V_{xyz} \in R^{3}$ in position $(x, y, z)$ is filled with the coordinates of its center under the egocentric camera coordinate system $(xL/N - L/2, yL/N - L/2, zL/N)$.

We project the 3D coordinates in $V$ into the egocentric image space with the fisheye camera model~\cite{scaramuzza2014omnidirectional}: $V_{\text{proj}} = \mathcal{P}(V)$, where $V_{\text{proj}}\in R^{N,N,N,2}$ and $\mathcal{P}$ is the fisheye camera projection function. The volumetric representation $V_{\text{body}}$ of the human body is obtained by filling a cube $~{V_{\text{body}} \in R^{N,N,N,K}}$ by bilinear sampling from the feature maps $F$ with $K$ channels using 2D coordinates in $V_{\text{proj}}$:
\begin{equation}
    V_{\text{body}} = F\{V_{\text{proj}}\}
\end{equation}
where $\{\cdot\}$ denotes bilinear sampling.

Then, we project the depth map to the 3D volumetric space. We first generate the point cloud of the scene geometry $C$ from the depth map $\hat{D}^S$ with the fisheye camera projection function $C = \mathcal{P}^{-1}(\hat{D}^S)$. The volumetric representation of scene depth map $V_{\text{scene}}$ is obtained by filling a binary cube $V_{\text{scene}} \in R^{N,N,N}$ by setting the voxel at $(x, y, z)$ to $1$ if there exists one point $(x_p, y_p, z_p)$ in the point cloud $C$ such that:
\begin{equation}
    \left\Vert (\frac{xL}{N} - \frac{L}{2}, \frac{yL}{N} - \frac{L}{2}, \frac{zL}{N}) - (x_p, y_p, z_p) \right\Vert < \epsilon
\end{equation}
where $\epsilon$ is the threshold distance.
In our experiment, we set $L = 2.4$ m, $N = 64$, and $\epsilon = 0.04$ m. This setting can cover most types of the human motions and allows high accuracy of the predicted body pose. 

\subsubsection{Predicting 3D Body Pose with V2V Network}

We feed the volumetric representation aggregated from $V_{\text{body}}$ and $V_{\text{scene}}$ into the volumetric convolutional network $\mathcal{N}$, which has a similar architecture as~\cite{moon2018v2v}. The V2V network produces the 3D heatmaps of the body joints:
\begin{equation}
    V_{\text{heatmap}} = \mathcal{N}(V_{\text{body}}, V_{\text{scene}})
\end{equation}

Following~\cite{iskakov2019learnable}, we compute the soft-argmax of $V_{\text{heatmap}}$ across the spatial axes to infer the body pose $\hat{P}$. The predicted pose $\hat{P}$ is finally compared with the ground truth pose $P_G$ from the EgoGTA dataset and $P_E$ from the EgoPW-Scene dataset with the MSE loss.




\section{Experiments} \label{experiments}
In this section, we evaluate our method considering existing and new datasets for egocentric monocular 3D human pose estimation. Please refer to the supplementary materials for the implementation details.


\begin{figure*}
\centering
\includegraphics[width=0.98\linewidth]{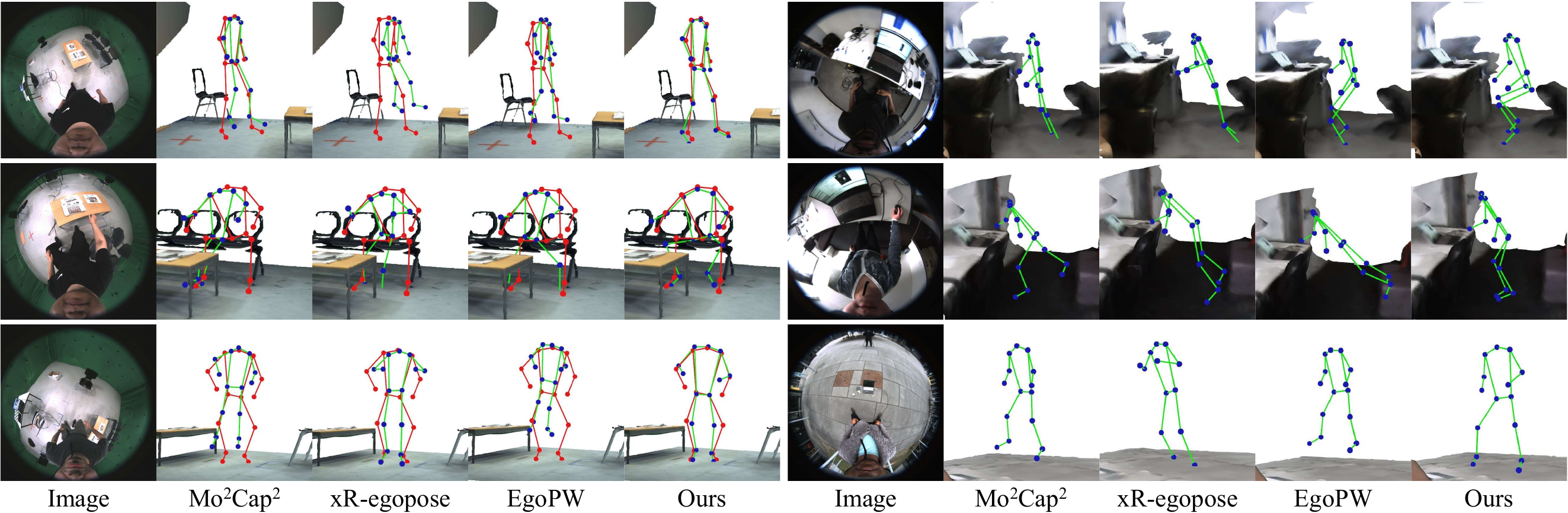}
\caption{Qualitative comparison between our method and the state-of-the-art egocentric pose estimation methods. From left to right: input image, Mo$^2$Cap$^2$ result, $x$R-egopose result, EgoPW result and our result. The ground truth pose is shown in red. The input images from the left part are from our test dataset, while those in the right part come from the EgoPW~\cite{Wang_2022_CVPR} in-the-wild test sequences (without ground-truth poses). We also show the gt scene geometry of the in-the-studio data and scene geometry obtained by SFM method for the in-the-wild data. For better visualizing the interaction between human body and environment, please refer to our supplementary video. \vspace{-1em}
}
\label{fig:results}
\end{figure*}

\subsection{Evaluation Datasets} \label{experiments:dataset_evaluation_metric}
%
%
Evaluating human-scene interaction requires precise annotations for camera pose and scene geometry. However, such information is not available in existing datasets for egocentric human pose estimation.
To solve this issue, we collected a new real-world dataset using a head-mounted fisheye camera combined with a calibration board.
The ground truth scene geometry is obtained with SfM method~\cite{hartley2003multiple} from a multi-view capture system with $120$ synced $4$K resolution cameras and the ground truth egocentric camera pose is obtained by localizing a calibration board rigidly attached to the egocentric camera.
This dataset contains around $28$K frames of two actors, performing various human-scene interacting motions such as sitting, reading newspaper, and using a computer. This dataset is evenly split into training and testing splits. We finetuned the method on the training split before the evaluation.
This dataset will be made publicly available and additional details of it are shown in the supplementary materials.

Besides our new test dataset, we also evaluate our methods in the test datasets from Wang~\etal~\cite{wang2021estimating} and Mo$^2$Cap$^2$~\cite{DBLP:journals/tvcg/XuCZRFST19}. The real-world dataset in Mo$^2$Cap$^2$~\cite{DBLP:journals/tvcg/XuCZRFST19} contains $2.7$K frames of two people captured in indoor and outdoor scenes, and the dataset in Wang~\etal~\cite{wang2021estimating} contains $12$K frames of two people captured in the studio.

\subsection{Evaluation Metrics}

We measure the accuracy of the estimated body pose with the MPJPE and PA-MPJPE. For the test dataset in Wang~\etal~\cite{wang2021estimating} and Mo$^2$Cap$^2$~\cite{DBLP:journals/tvcg/XuCZRFST19}, we evaluate PA-MPJPE and BA-MPJPE~\cite{DBLP:journals/tvcg/XuCZRFST19} since the ground truth poses in the egocentric camera space are not provided. Further details of the metrics are shown in the supplementary materials.

\subsection{Comparisons on 3D Pose Estimation} \label{experiments:main_compare}

\begin{table}[h]
\begin{center}
\small
\begin{tabularx}{0.47\textwidth} { 
   >{\raggedright\arraybackslash}X 
   >{\centering\arraybackslash}X 
   >{\centering\arraybackslash}X  }
\hlineB{2.5}
Method & MPJPE & PA-MPJPE \\
\hline
\multicolumn{2}{l}{\textbf{Our test dataset}} \\
\hline
Mo$^2$Cap$^2$~\cite{DBLP:journals/tvcg/XuCZRFST19} & 200.3 & 121.2 \\
$x$R-egopose~\cite{DBLP:conf/iccv/TomePAB19} & 241.3 & 133.9 \\
EgoPW~\cite{Wang_2022_CVPR} & 189.6 & 105.3 \\
Ours & \textbf{118.5} & \textbf{92.75} \\
\hlineB{2.5}
Method & PA-MPJPE & BA-MPJPE \\
\hline
\multicolumn{2}{l}{\textbf{Wang~\etal's dataset~\cite{wang2021estimating}}} \\
\hline
Mo$^2$Cap$^2$~\cite{DBLP:journals/tvcg/XuCZRFST19} & 102.3 & 74.46 \\
$x$R-egopose~\cite{DBLP:conf/iccv/TomePAB19} & 112.0 & 87.20 \\
EgoPW~\cite{Wang_2022_CVPR} & 81.71 & 64.87 \\
Ours & \textbf{76.50} & \textbf{61.92} \\
\hlineB{2.5}
\multicolumn{2}{l}{\textbf{Mo$^2$Cap$^2$ test dataset~\cite{DBLP:journals/tvcg/XuCZRFST19}}} \\
\hline
Mo$^2$Cap$^2$~\cite{DBLP:journals/tvcg/XuCZRFST19} & 91.16 & 70.75 \\
$x$R-egopose~\cite{DBLP:conf/iccv/TomePAB19} & 86.85 & 66.54 \\
EgoPW~\cite{Wang_2022_CVPR} & 83.17 & 64.33 \\
Ours & \textbf{79.65} & \textbf{62.82} \\
\hlineB{2.5}
\end{tabularx}
\end{center}
\caption{Performance of our method on our test dataset, Wang~\etal's test dataset~\cite{wang2021estimating} and Mo$^2$Cap$^2$ test dataset~\cite{DBLP:journals/tvcg/XuCZRFST19}. Our method outperforms the state-of-the-art methods EgoPW,  Mo$^2$Cap$^2$ \cite{DBLP:journals/tvcg/XuCZRFST19} and $x$R-egopose \cite{DBLP:conf/iccv/TomePAB19}.
\vspace{-1em}
}
\label{table:main}
\end{table}

In this section, we compare 
our approach
with previous single-frame-based methods, including EgoPW~\cite{Wang_2022_CVPR}, $x$R-egopose~\cite{DBLP:conf/iccv/TomePAB19} and Mo$^2$Cap$^2$~\cite{DBLP:journals/tvcg/XuCZRFST19}
on our test dataset under the ``Our test data''
in Table~\ref{table:main}. Since the code for $x$R-egopose is not released, we use our implementation for the evaluation. In our dataset, the proposed method outperforms the previous state-of-the-art methods, EgoPW~\cite{Wang_2022_CVPR} by 37.5\% on MPJPE and 11.9\% on PA-MPJPE.
We also compared with previous methods on the Wang~\etal's test dataset~\cite{wang2021estimating} and Mo$^2$Cap$^2$ test dataset~\cite{DBLP:journals/tvcg/XuCZRFST19} and show the results in Table~\ref{table:main}.
On Wang~\etal's test dataset, our method performs better than EgoPW by 6.4\%. On the Mo$^2$Cap$^2$ test dataset, our method performs better than EgoPW by 7.8\%.


\begin{table}
\begin{center}
\small
\begin{tabularx}{0.47\textwidth} { 
   >{\raggedright\arraybackslash}X 
   >{\centering\arraybackslash}X 
   >{\centering\arraybackslash}X  }
\hlineB{2.5}
Method & Non pene. & Contact \\
\hline
Mo$^2$Cap$^2$~\cite{DBLP:journals/tvcg/XuCZRFST19} & 69.6\% & 23.1\% \\ 
$x$R-egopose~\cite{DBLP:conf/iccv/TomePAB19} & 64.5\% & 38.3\% \\  
EgoPW~\cite{Wang_2022_CVPR} & 71.7\% & 38.8\% \\  
Ours & \textbf{84.1\%} & \textbf{89.4\%} \\  
\hlineB{2.5}
\end{tabularx}
\end{center}
\caption{
Comparisons of physical plausibility on our test dataset. 
}
\label{table:interaction}
\end{table}

We also evaluate the physical plausibility of our predictions by calculating the percentage of predicted poses that are in contact with the scene and do not penetrate the scene in Table~\ref{table:interaction}. We define a body pose as in contact with the scene if any body joint is less than $5$cm from the scene mesh. A body pose suffers from the body floating issue if it is not in contact with the scene.
Compared with previous approaches, our method generates body poses more physically plausible considering the constraints of the scene.

From the results in Table~\ref{table:main} and Table~\ref{table:interaction}, we can see that our approach outperforms all previous methods on the single-frame egocentric pose estimation task. For the qualitative comparison, we show the results of our method on the studio dataset and in-the-wild sequences in Fig.~\ref{fig:results}. See our supplementary video for more qualitative evaluations.
Our predicted poses are physically plausible under the scene constraint, whereas other methods generate poses suffering from body floating and penetration issues. 
%

\begin{figure}
\centering
\includegraphics[width=0.96\linewidth]{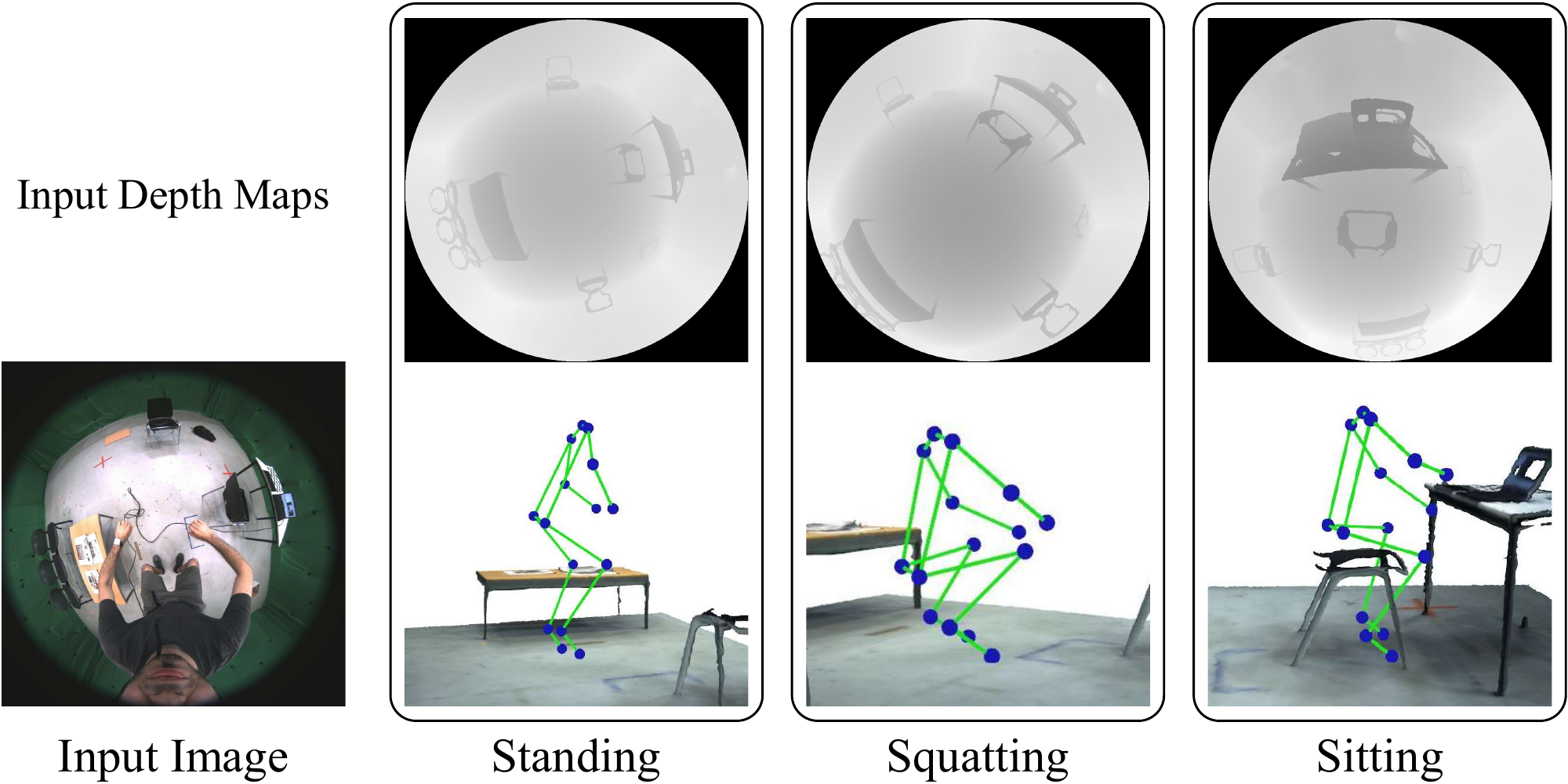}
\caption{Predicted pose with different scene depth map input. Our network can generate different poses under different depth input and further disambiguate body poses under scene constraint.
}
\label{fig:change_depth}
\end{figure}

In order to further demonstrate our method performing predictions accordingly to the constraint of the scene geometry, we fix the input image and change the scene depth input to the depth map corresponding to the standing pose, squatting pose, and sitting pose. The results are presented in Fig.~\ref{fig:change_depth} and show that the predicted poses change to standing, squatting, and sitting to better adapt to the input changes of the scene geometry. 
This shows our method's ability to disambiguate poses under different scene constraints.

\subsection{Ablation Study}\label{experiments:ablation}

\paragraph{Simple Combination of 2D Features and Depth Maps.}
In Sec.~\ref{method:pose}, we claim that the volumetric representation of egocentric 2D features and scene depth map is important for understanding the interaction between the human body and the surrounding scene. To provide evidence of this claim, in this experiment, we compare our method with baseline methods that simply combine the 2D image features and scene depth map. We set two baseline methods since there are two types of egocentric pose estimation methods, \ie direct regression of 3D poses ($x$R-egopose~\cite{DBLP:conf/iccv/TomePAB19}) and prediction of 2D position and depth for each joint (Mo$^2$Cap$^2$~\cite{DBLP:journals/tvcg/XuCZRFST19} and EgoPW~\cite{Wang_2022_CVPR}). In the baseline method ``$x$R-egopose + Depth'', we concatenate 2D heatmaps and scene depth map as the input to the 3D pose regression network in $x$R-egopose. In the baseline ``EgoPW + Depth'', we concatenate the 2D features and the scene depth map and input them into the joint depth prediction network.

From the evaluation results shown in Table~\ref{table:ablation}, both of the baseline methods perform worse than our proposed method. In ``$x$R-egopose + Depth'', simply combining the scene depth and 2D heatmaps cannot provide direct geometric supervision for the 3D pose regression network. In ``EgoPW + Depth'', though the joint depth estimation network performs better with the help of scene depth information, the 2D pose estimation network does not benefit from it. Both of the experiments demonstrate the effectiveness of our volumetric representation of 2D features and scene geometry, which provides direct geometry supervision for the physical plausibility of predicted egocentric poses.

\begin{table}
\begin{center}
\small
\begin{tabular}{p{0.2\textwidth}
>{\centering}p{0.1\textwidth}
>{\centering\arraybackslash}p{0.1\textwidth}}
\hlineB{2.5}
Method & MPJPE & PA-MPJPE \\
\hline
EgoPW+Optimizer & 187.1 & 103.2 \\
EgoPW+Depth & 149.6 & 98.15 \\
\hline
$x$R-egopose+Depth & 180.5 & 103.7 \\
\hline
Ours w/o Depth & 188.1 & 105.1 \\
Ours+Depth with Body & 167.3 & 103.3 \\
Ours+Depth w/o Body  & 135.7 & 95.84 \\
Ours+Depth w/o Inpainting  & 124.2 & 95.00 \\
Ours+GT Depth & \textbf{109.9} & \textbf{88.80} \\
\hline
Ours & \textit{118.5} & \textit{92.75} \\
\hlineB{2.5}
\end{tabular}

\end{center}
\caption{Results from our method compared to different baselines.
}
\label{table:ablation}
\end{table}

\paragraph{Optimization.}
In this experiment, we compare our method with an optimization baseline that refines a 3D pose considering the scene constraint. Similar to Wang~\etal~\cite{wang2021estimating} and EgoPW~\cite{Wang_2022_CVPR}, we first train a VAE consisting of a CNN-based encoder $f_{\text{enc}}$ and decoder $f_{\text{dec}}$ to learn an egocentric motion prior. Then, We optimize the egocentric pose $P$ by finding a latent vector $z$ such that the corresponding pose $P = f_{\text{dec}}(z)$ minimizes the objective function $E(P) = \lambda_R E_R + \lambda_J E_J + \lambda_C E_C$,
where $E_R$ is the egocentric reprojection term, $E_J$ is the egocentric pose regularization term, and $E_C$ is the contact term. The latent vector $z$ is initialized with the estimated pose from the EgoPW method. The $E_R$ and $E_J$ are the same as those defined in Wang~\etal~\cite{wang2021estimating}. Denote the $n$th joint in egocentric pose $P$ as $P_n, n\in [0, N]$, where $N$ is the number of joints, and the $m$th point in scene point cloud $C$ as $C_m, m\in [0, M]$, where $M$ is the number of points in the point cloud. The contact term $E_C$ is defined as:
\begin{equation}
\begin{split}
    E_C &= \sum_{n\in [0, N]} d_n^2,\ \quad \text{if } d_n \le \epsilon, \ \text{otherwise}\ 0,\; \text{and} \\
    d_n & = \min_{m\in [0, M]} \Vert P_n - C_m \Vert_2.
\end{split}
\end{equation}
We first calculate the nearest distance $d_n$ between each body joint and the projected point cloud $C$ from the scene depth map. If the distance $d_n$ of the $n$th joint is less than a margin $\epsilon$, it is defined as in contact with the scene and minimized with the optimization framework. 

The result of the optimization method is shown as ``EgoPW+Optimizer'' in Table~\ref{table:ablation}, which demonstrates that the optimization framework is less effective than our method. This is because the accuracy of optimization method relies heavily on the initial pose. If the initial pose is not accurate, it will be difficult to determine the contact labels for each joint with the fixed distance margin. Without accurate contact labels, the optimization framework might force the joint that does not contact the scene to keep in contact, eventually resulting in wrong poses.

\paragraph{Scene Depth Estimator.}

In our work, we estimate the depth of the surrounding scene and infer the depth behind the human body with a depth inpainting network. In order to validate the effectiveness of the scene depth map, we remove the input depth map from the V2V network and show the results of ``Ours w/o Depth'' in Table~\ref{table:ablation}. This baseline increases the MPJPE by about $70$ mm, which is evidence of the relevant extra information provided by the scene depth.

To demonstrate the benefits of recovering the scene depth behind the human body, we evaluate our model using the estimated depth including the human body as the input to the V2V network. We also removed the human body area from the depth
maps and use them as input to the V2V network. The results are shown in ``Ours+Depth with Body'' and ``Ours+Depth w/o Body'' in Table~\ref{table:ablation}.
Both of the baseline methods perform worse than our method because the area in the scene occluded by the human body can provide clues for generating plausible poses.

We also evaluate our method with ground truth depth maps in ``Ours+GT Depth'' in Table~\ref{table:ablation}, which further improves over our estimated depth by 7.2\% in terms of MPJPE and 4.2\% in terms of PA-MPJPE. This demonstrates that the accuracy of our predicted pose benefits from better depth maps, but still our estimated scene depth already provides a significant improvement over the baselines. 

\begin{figure}
\centering
\includegraphics[width=0.96\linewidth]{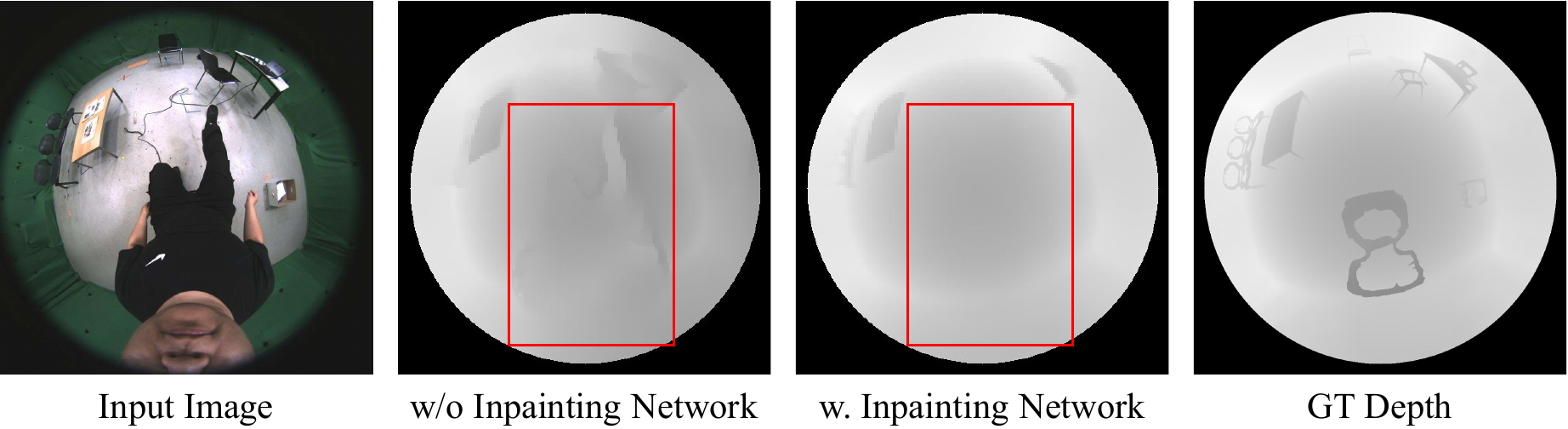}
\caption{
The qualitative depth estimation results with or without inpainting network. The depth map estimated without inpainting network show artifacts in the human body region (see the red box).
}
\label{fig:depth}
\end{figure}

Finally, we compare with a baseline method without depth inpainting, \textit{i.e.}, the human body is removed from the input image and the scene depth map is predicted directly with the network $\mathcal{D}$ in Sec.~\ref{method:depth}. The pose estimation accuracy is shown in ``Ours+Depth w/o Inpainting'' of Table~\ref{table:ablation} and the depth accuracy is shown in Table~\ref{table:depth}.
Our method outperforms the baseline as it is more challenging to simultaneously estimate and inpaint the depth. Moreover, the network can be influenced by the segmented part in the input image as some extinct object, as shown in Fig~\ref{fig:depth}. 
\begin{table}[h]
\begin{center}
\small
\begin{tabular}{p{0.2\textwidth}
>{\centering}p{0.1\textwidth}
>{\centering\arraybackslash}p{0.1\textwidth}}
\hlineB{2.5}
Method & Abs-Rel & RMSE(m) \\
\hline
w/o Inpainting Network  & 0.3834 & 0.7856 \\
Ours & 0.1069 & 0.3365 \\
\hlineB{2.5}
\end{tabular}
\end{center}
\caption{
The quantitative results of depth estimator. We evaluate Abs-Rel and RMSE on our test dataset following~\cite{hu2019revisiting}.
\vspace{-1em}
}
\label{table:depth}
\end{table}

\section{Conclusion} \label{sec:conclusion}
In this paper, we have proposed a new approach to estimate egocentric human pose under the scene constraint. We firstly train a depth inpainting network for estimating the depth map of the scene without human body. Next, we combine the egocentric 2D features and scene depth map in a volumetric space and predict the egocentric pose with a V2V network.
The experiments show that our method outperforms all of the baseline methods both qualitatively and quantitatively and our method can predict physically plausible poses in terms of human-scene interaction. 
In future, this method could be extended to estimate physically-plausible egocentric motion from a temporal sequence. 

\noindent\textbf{Limitations.} The accuracy of voxel-based pose estimation network is constrained by the accuracy of estimated depth, especially where the scene is occluded by the human body. One solution is to leverage the temporal information to get a full view of the surrounding environment.

\noindent \textbf{Acknowledgments} Jian Wang, Diogo Luvizon, Lingjie Liu, and Christian Theobalt have been supported by the ERC Consolidator Grant 4DReply (770784).

{\small
\bibliographystyle{ieee_fullname}
\bibliography{egbib}
}

\end{document}